\pdfoutput=1

\documentclass[11pt]{article}

\usepackage[preprint]{coling}

\usepackage{times}
\usepackage{latexsym}

\usepackage[T1]{fontenc}

\usepackage[utf8]{inputenc}

\usepackage{microtype}

\usepackage{inconsolata}

\usepackage{graphicx}

\usepackage[acronym]{glossaries}
\glsdisablehyper
\usepackage{amsfonts,amsmath,amssymb}
\usepackage{todonotes}
\usepackage{svg}
\usepackage{graphicx}
\usepackage{pifont}

\usepackage{xcolor}
\usepackage{xstring}
\usepackage[table-align-uncertainty, separate-uncertainty=true]{siunitx}
\usepackage[noabbrev]{cleveref}
\usepackage{svg}
\usepackage{tabularx}
\usepackage{enumitem}
\usepackage{footnote}
\makesavenoteenv{tabular}
\makesavenoteenv{table}
\usepackage{tablefootnote}
\usepackage{adjustbox}

\usepackage{listings}
\newcommand{\mask}{\texttt{MASK}}
\newcommand{\T}{T}
\newcommand{\M}{\mathcal M}
\newcommand{\x}{\ensuremath{\mathbf x} }
\newcommand{\Dl}{\mathcal{D}}
\newcommand{\Dtr}{\mathcal{D}_\text{train}}
\newcommand{\Dv}{\mathcal{D}_\text{valid}}

\newcommand{\cmark}{\ding{51}}%
\newcommand{\xmark}{\ding{55}}%

\definecolor{C0}{HTML}{a1c9f4}
\definecolor{C1}{HTML}{ffb482}
\definecolor{C2}{HTML}{8de5a1}
\definecolor{C3}{HTML}{ff9f9b}
\definecolor{C4}{HTML}{8172b3}
\definecolor{C5}{HTML}{937860}
\definecolor{C6}{HTML}{da8bc3}

\title{Manual Verbalizer Enrichment for Few-Shot Text Classification}

\author{Quang Anh Nguyen$^{\ast \ddag}$, Nadi Tomeh$^{\dag}$, Mustapha Lebbah$^{\ast}$, \\ \bf\large Thierry Charnois$^{\dag}$, Hanene Azzag$^{\dag}$, Santiago Cordoba Muñoz$^{\ddag}$ \\
$^{\ast}${Université Paris-Saclay - DAVID Lab, UVSQ, Versailles, France} \\
$^{\dag}${Universit\'e Sorbonne Paris Nord - LIPN CNRS UMR 7030, Villetaneuse, France} \\
$^{\ddag}${Groupe BPCE - Paris, France}} 

\begin{document}
\maketitle
\begin{abstract}
With the continuous development of pre-trained language models, prompt-based training becomes a well-adopted paradigm that drastically improves the exploitation of models for many natural language processing tasks. Prompting also shows great performance compared to traditional fine-tuning when adapted to zero-shot or few-shot scenarios where the number of annotated data is limited. In this framework, the role of verbalizers is essential, as an interpretation from masked word distributions into output predictions. In this work, we propose MaVEN, an approach for verbalizer construction by enrichment of class labels using neighborhood relation in the embedding space of words for the text classification task. In addition, we elaborate a benchmarking procedure to evaluate typical baselines of verbalizers for document classification in few-shot learning contexts. Our model achieves state-of-the-art results while using significantly fewer resources. We show that our approach is particularly effective in cases with extremely limited supervision data. Our code is available at \url{https://anonymous.4open.science/r/verbalizer_benchmark-66E6}.
\end{abstract}

\section{Introduction}

\label{sec:introduction}

\newacronym{plm}{PLM}{pre-trained language model}
\newacronym{nlp}{NLP}{natural language processing}
\newacronym{lm}{LM}{language model}

Fine-tuning PLM \citep{devlin2019bert,zhuang-etal-2021-robustly,brown2020language} resulted in large improvements in various \acrshort{nlp} tasks. Classic approaches replace the \acrshort{plm}'s output layer with a task-specific head and fine-tune the entire model \cite{devlin2019bert, liu2019roberta, JMLR:v21:20-074}. However, additional classification layers import a great amount of randomly initialized parameters that need a sufficient amount of labeled data to be trained. Classical fine-tuning, therefore becomes inapplicable for few-shot or zero-shot scenarios \cite{yin-etal-2019-benchmarking, zhang-etal-2023-revisit}. 

Prompting has become a novel paradigm where downstream tasks are transformed to suit the pre-training objective.
Prompt-based fine-tuning allows to exploit \acrshortpl{plm}' knowledge while reducing the gap between pre-training and fine-tuning \cite{petroni-etal-2019-language, Chen_2022}.
In this framework, templates and verbalizers \cite{schick-schutze-2021-exploiting, gao-etal-2021-making} are crucial elements to map between task-specific inputs and labels, to textual data for the \acrshort{lm}.  For example, given a piece of text:
\begin{align*}
    \x=\text{``Dollar rises against euro...''}
\end{align*}
The task is to predict if this text belongs to which class among politics, sports, science, or economics. A \textit{template} $T$ first transforms the given text into a cloze question. For instance, one may choose for this task: 
\begin{align*}
    T(\x) = \text{``\_\_\_ news: Dollar rises against euro...''}
\end{align*}

The task of predicting labels without conceptual meaning is transformed into identifying whether the most probable choice for the masked position \_\_\_ is ``politics'', ``sports'', ``science'' or ``economics''. This task, namely masked language modeling aligns coherently with the pre-training of a variety of masked LMs, notebly BERT \cite{devlin-etal-2019-bert}, RoBERTa \cite{zhuang-etal-2021-robustly}.
 
A masked LM takes the wrapped text, marks the missing position with its \mask{} token, and produces probabilities for the masked token over the vocabulary. Ideally in this case, one would expect the probability of the word ``economics'' to be higher than that of ``sports''. This straightforward approach maps each class to a single word, its textual name. In general, a \textit{verbalizer} refers to a mapping from the label space to the vocabulary space, where each label is mapped to multiple vocabulary tokens.


\newacronym{mave}{MaVEN}{Manual Verbalizer Enrichment by Nearest Neighbors’ Embeddings}

\newacronym{npp}{NPPrompt}{Nonparametric Prompting}

In many cases, verbalizers are defined \textit{manually} using human knowledge of the downstream task, to choose words that semantically represent the meaning of class labels \cite{schick-schutze-2021-exploiting, schick-schutze-2021-just, gao-etal-2021-making}. There exists other constructions such as \textit{soft} verbalizers \cite{hambardzumyan-etal-2021-warp, cui-etal-2022-prototypical}.  Algorithms for \textit{automatic} label word searching exist in the literature. One such example is PETAL \cite{schick-etal-2020-automatically}, where label words are mined based on their likelihood on supervised data. We notice that the procedure presented in \cite{schick-schutze-2021-exploiting, schick-etal-2020-automatically} includes semi-supervised learning and therefore additional unlabeled data. One another example is KPT \cite{hu-etal-2022-knowledgeable} where an external knowledge base such as WordNet \cite{miller-1994-wordnet} and ConceptNet \cite{speer-havasi-2012-representing} are used to expand label words from the class name. Our motivation in this work is to propose a method to enrich the manual verbalizer without resorting to external resources. Among various techniques, \acrfull{npp} \cite{zhao-etal-2023-pre} uses PLM’s embeddings to find relevant words to labels automatically. However, NPPrompt is designed exclusively for zero-shot learning and presents many shortcomings (see \cref{sub:maven}), thus our motivation to develop this idea for few-shot learning by enrichment of manual verbalizers. In this paper, we also do an extensive ablation study on the effect of multiple elements of the proposed algorithm on verbalization performance in few-shot text classification.



Our contribution is summarized as follows:
\begin{enumerate}[label=(\roman*)]
    \item
    We propose an extended formulation of \acrshort{npp} to enrich the manual verbalizer by neighbors in the embedding space for few-shot finetuning, which achieves improved performance over previous work, particularly with an extremely limited amount of data.
    \item
    In a template-independent manner, we systematically compare this method to manual, soft, and automatic verbalizers for the text classification task. The results are presented on three English public datasets previously studied in the literature. We also present new results on two French datasets.
    \item 
    We conduct ablation experiments on multiple elements of the proposed algorithm.
\end{enumerate}

\section{Related Works}
\label{sec:background}

\paragraph{Prompt-based fine-tuning} In this framework, the input is wrapped with a task-specific \emph{template} to reformulate the classification task as language modeling as described in \cref{sec:introduction}. The \emph{verbalizer} then transforms the distribution of the \mask{} token into label prediction (see \cref{sec:methodology} for formal definitions). The choice of textual templates and verbalizer, have a significant influence on the classification performance \cite{gao-etal-2021-making}.

PET and iPET \cite{schick-schutze-2021-exploiting, schick-schutze-2021-just} use task-specific manual templates and verbalizers that work efficiently. 
However, their construction requires both domain expertise of downstream tasks and understanding of biases in the \mask{} distribution produced by the \acrshortpl{plm}. Otherwise, the search process for an optimal template and verbalizers may be computationally exhaustive with a large number of classes. Meanwhile, \cite{lester-etal-2021-power, liu-etal-2022-p, li-liang-2021-prefix} propose to freeze the \acrshort{plm} and instead optimize prompt tokens. Despite being human-independent and storage-saving, continuous prompts have only been studied in data-abundant scenarios, and produce tokens that are hard to interpret. Here, we study textual templates instead and focus on the search for label words for the verbalizer. 


\paragraph{Enrichment of manual verbalizer} Previous works also propose methods to improve the semantics of label words for a given manual verbalizer. KPT \cite{hu-etal-2022-knowledgeable} incorporates external knowledge into the verbalizers, along with multiple steps of refinement and calibration to obtain words with wide coverage of given classes.
Still, such knowledge bases may not always be available. Therefore, we are motivated to derive a method to improve the manual verbalizer independently from additional resources.
On the other hand, NPPrompt \cite{zhao-etal-2023-pre} searches for cognates of initial manual words using the embedding layer of the same PLM. This approach attains greater coherence in later PLM fine-tuning.

\section{Methodology}
\label{sec:methodology}

Let $\M$ be a language model with vocabulary $V$. Following \cite{schick-schutze-2021-exploiting, schick-schutze-2021-just}, we define the template - verbalizer pair. Let $(\x, y)$ be an example of the classification problem, where $\x$ represents one or many sentences and $y$ is its label in the label set $\mathcal Y$. A template $T$ maps $\x$ into a masked sequence $T(\x)$ of tokens in $V \cup \{ \mask\}$. A verbalizer $v: \mathcal Y \rightarrow \mathcal P(V)$ maps each label to a set of words characterizing the class (called label words). The probability of the label conditioned on the input is then modeled by the logits of its label words conditioned on the masked sequence:
\begin{align}
    \label{eq:1}
    p(y\vert \x) \propto {\exp \left(\frac{1}{\left\vert v(y)\right\vert} \sum_{w \in v(y)}  \M \left( w \middle\vert \T(\x) \right)\right)}
\end{align}
Where $\mathcal M (w \vert T(\x))$ denotes the logit of \mask\  being predicted as $w$ by the LM conditional on the masked sequence $T(\x)$. 

\subsection{Baselines}
\label{sub:baselines}
\paragraph{Manual} Label words can be predefined manually from users' knowledge of classes \cite{gao-etal-2021-making,schick-schutze-2021-exploiting}. To minimize the necessity of domain expertise, here the manual verbalizers derive directly from the class names.

\paragraph{Soft} WARP \cite{hambardzumyan-etal-2021-warp} proposes to represent each label $y$ by a prototype vector $v_y$ instead of concrete words, initialized with static embeddings of the manual label words and optimized alongside the \acrshort{plm}.

\paragraph{Auto} Among automatic methods, PETAL \cite{schick-etal-2020-automatically} allows identifying words suitable to represent classes from training data. Consider the classification problem as many one-vs-rest binary problems to find label words for each class separately. For each label, PETAL takes the top $k_{\rm auto}$ words that maximize the likelihood ratio of positive examples and minimize that of negative examples.

\newacronym{llm}{LLM}{large language model}

In addition to applying verbalizers to small masked \acrshortpl{lm}, we also evaluate the performance \acrfullpl{llm} as follows.
\newacronym{it}{Instruct}{instruction tuning}

\paragraph{Instruction tuned LLM (Instruct)} Instruction tuning
is an effective technique to enhance the capabilities and controllability of \acrshortpl{llm} \cite{zhang2024instruction,wei2022finetuned}. It involves further
training of the generative \acrshortpl{llm} using textual (instruction, output) pairs. Numerous instruction-tuned \acrshortpl{llm}, including InstructGPT \cite{ouyang2022training}, Flan-T5 \cite{chung2022scaling}, T0 \cite{sanh2022multitask}, BLOOMZ \cite{muennighoff2023crosslingual}, etc. achieve remarkable zero-shot performance. They mainly differ in their backbone model and their instruction dataset construction.

\newacronym{icl}{ICL}{in-context learning}

We use \texttt{Mistral-7B-Instruct-v0.2}, an instruction-tuned version of \texttt{Mistral-7B-v0.2} \cite{jiang2023mistral}, for its reasonable size. \texttt{Mistral} is publicly available and  achieves state-of-the-art performance compared to similar-sized \acrshortpl{llm}. 
The prompt is adapted from P3 \cite{bach2022promptsource} for zero-shot inference. For few-shot inference, \cite{dong2023survey}, \textbf{\acrfull{icl}} is combined with the instruction, where labeled examples are included in the prompt as a demonstration. Due to machine memory limitations, we only apply \acrshort{icl} for $N=32$. See \cref{app:mistral} for specific prompts.

\subsection{Manual Verbalizer Enrichment by Nearest Neighbors' Embeddings} 
\label{sub:maven}

\newacronym{mave}{MaVEN}{\textbf{Ma}nual \textbf{V}erbalizer \textbf{E}nrichment by Nearest \textbf{N}eighbors' Embeddings}

In this paper, we propose \acrfull{mave}, an extended formulation of \acrshortpl{npp} \cite{zhao-etal-2023-pre}, adapted for prompt-based finetuning. Noting that the probability score that the \acrshort{lm} assigns to a specific topic is dispersed over multiple label words, we hypothesize that the manual verbalizer captures only a part of this mass and thus is sensitive to the choice of label words. Our motivation therefore is to automatically extend the verbalizer to capture more semantic information by including semantically related words.

In most practical scenarios, a natural manual verbalizer can often be obtained using the names of classes, as class names naturally encode the semantic meaning of texts belonging to the class. We assume that for our classification problem, let $v$ be the initial manual verbalizer. In our case, $v(y)$ includes words extracted directly from the name of the class $y$. Let $E$ be a word embedding function, the word embedding layer of the LM for example. For each core word $w_0\in v(y)$, we define the neighborhood of $w_0$ as:
\begin{align}
    \mathcal N_k(w_0) = \{w_0\} \cup \underset{w}{\mathrm{top-} k} \left[s \left( {w_0, w} \right) \right]
\end{align}
Where $s$ is the cosine similarity in this embedding space $E$.

We enlarge the verbalizer $v(y)$ as the union of neighborhoods of all core words:
\begin{align}
    \hat v(y) = \bigcup_{w_0 \in v(y)} \mathcal N_k (w_0)
\end{align}

The hyperparameter $k$ represents the size of the neighborhood in the embedding space around the initial core words. In our experiments, without specifying differently, we take $k=15$.

The probability of the class $y$ is aggregated over its augmented verbalizer as follows:
\begin{align}
    p (y\vert \x) \propto \exp \left( \frac{\sum_{w \in \hat v(y)} q^y_w \mathcal M(w \vert T(\x))}{\sum_{w \in \hat v(y)} q^y_w } \right)
\end{align}

The weights $q^y_w$ represent the contribution of the word $w \in \mathcal N_k(w_0)$ in the class $y$. 

In comparison with the original method \acrshort{npp} \cite{zhao-etal-2023-pre}, which focuses exclusively in zero-shot setting, our work differs in many adaptations for finetuning:
\begin{enumerate}[label=$\bullet$]
    \item 
    Neighborhood-level aggregation: Each $q^y_w$ is initialized by the similarity $s({w, w_0})$ of $w$ to its core word $w_0$ and fine-tuned with the parameters of the PLM. 
    \item
    Class-level aggregation: if a class is represented by more than one  (meaning that $v(y)$ contains multiple core words), instead of taking the neighborhood with highest score as \cite{zhao-etal-2023-pre}, we merge the neighborhoods and calculate the class score from all merged neighbors. This way the aggregated class score is a derivable the function of the \acrshort{plm} outputs.
    \item 
    Template selection: While \cite{zhao-etal-2023-pre} reports the result of the best template (on the test set itself), we find this process unjust and biased. To avoid cherry-picking and reduce template dependence, we follow the ensemble aproach detailed in the following paragraph.  
\end{enumerate}

After identifying the label words, the \acrshortpl{plm} are fine-tuned based on the chosen template and verbalizer, by minimizing the cross entropy loss between the predicted probabilities and the correct labels. Given the sensitivity of prompt-based methods in a few-shot context, each prompt can more or less effectively elicit knowledge from the \acrshort{plm}. The ensemble approach provides an efficient way to reduce instability across prompts and provide stronger classifiers \cite{schick-schutze-2021-exploiting, jiang-etal-2020-know}. We also study the impact of aggregating strategy. The logits of individual models trained on different templates are aggregated into the final prediction, using three aggregation strategies: (vote) majority vote from individual predictions, (proba) averaging individual class probabilities, and (logit) averaging individual class logits.

\section{Experiments}

\subsection{Settings}
\label{sec:experiment}
Five datasets (\cref{sec:data}) are considered context for three baselines (\cref{sec:methodology}) and \acrshort{mave} in few-shot prompt-based fine-tuning. For each dataset, from the original training set, we sample a labeled set $\Dl$ of cardinality $N$. For each run, split $\Dl$ into two halves: $\Dtr$ is used for fine-tuning with the template - verbalizer pair and $\Dv$ for validation \cite{zheng-etal-2022-fewnlu}. The best checkpoint is retained from the score obtained on the validation set. Details of hyperparameters is in \cref{app:params}. 

The underlying \acrfull{plm} is \texttt{RoBERTa-large} \cite{liu2019roberta} as in \cite{schick-etal-2020-automatically} for datasets in English, or \texttt{CamemBERT-large} \cite{martin-etal-2020-camembert} for datasets in French. We report the average and standard deviation of accuracy from 3 repetitions with different samplings of $\Dl$, to evaluate the result variation with different training data instances. 

Our implementation is based on the toolkit OpenPrompt \cite{ding-etal-2022-openprompt} and the Transformers package \cite{wolf-etal-2020-transformers}. Experiments are executed on two types of GPUs: NVIDIA Tesla V100 and NVIDIA Quadro RTX 5000. 

\subsection{Datasets and templates}
\label{sec:data}
Our experiments are done on three public English datasets and two datasets in French (\cref{tab:data}). For each dataset, four textual templates are created. The manual verbalizers for each dataset can be found in \cref{app:mv}.

\begin{table}[h]
    \centering
    \begin{tabular}{lcccc}
    \hline
        Dataset     & Classes       & Test set      & Balanced      \\
    \hline
        AG          & 4             & 7600          & \cmark        \\
        DBpedia     & 14            & 75000         & \cmark        \\
        Yahoo       & 10            & 60000         & \cmark        \\
        FrN         & 10            & 536           & \xmark        \\
        MLSUM Fr    & 10            & 10585         & \xmark        \\
    \hline
    \end{tabular}
    \caption{Dataset details.}
    \label{tab:data}
\end{table}

\paragraph{AG} AG's News \citep{NIPS2015_250cf8b5}. Given a headline \x, a news needs to be classified into one of 4 categories. For this dataset:

    \begin{align*}
        \T_0(\x) &= {\mask} \text{ news: } \x
        \\
        \T_1(\x) &= \x \text{ This topic is about \mask.}
        \\
        \T_2(\x) &= \text{[Category: \mask] } \x
        \\
        \T_3(\x) &= \text{[Topic: \mask] } \x
    \end{align*}

\paragraph{DBpedia} The DBpedia ontology classification dataset \cite{NIPS2015_250cf8b5} is constructed by picking 14 non-overlapping classes from DBpedia 2014. Given a title $\x_1$ and its description $\x_2$, the task is to predict the category of the object in the title.
    \begin{align*}
        \T_0(\x) &= \text{$\x_1 \x_2$ In this sentence, $\x_1$ is \mask.}
        \\
        \T_1(\x) &= \text{$\x_1 \x_2$ $\x_1$ is \mask.}
        \\
        \T_2(\x) &= \text{$\x_1 \x_2$ The category of $\x_1$ is \mask.}
        \\
        \T_3(\x) &= \text{$\x_1 \x_2$ The type of $\x_1$ is \mask.}
    \end{align*}

\paragraph{Yahoo} Yahoo! Answers \citep{NIPS2015_250cf8b5} is
a text classification dataset of questions from Yahoo!. Given a question (title and content) and its answer, one of ten possible categories has to be assigned.
For a concatenation \x of the question title, question content and the answer, we define: 
    \begin{align*}
        \T_0(\x) &= \mask \text{ question: $\x$.} 
        \\
        \T_1(\x) &= \x \text{ This topic is about \mask.} 
        \\
        \T_2(\x) &= \text{[Topic: \mask] $\x$.} 
        \\
        \T_3(\x) &= \text{[Category: \mask] $\x$.} 
    \end{align*}

\paragraph{MLSUM Fr} originated from MultiLingual SUMmarization \cite{scialom-etal-2020-mlsum}, a large-scale dataset from online newspapers. From this base, the French split is preprocessed and annotated for the task of topic classification by grouping the topic tag into one of ten categories\footnote{We follow the procedure presented by reciTAL teams at \url{https://huggingface.co/lincoln/flaubert-mlsum-topic-classification}.}.

\newacronym{frn}{FrN}{French News}
\paragraph{\acrshort{frn}} This real-world private dataset in French is provided by our collaborator in a private company, consisting of press articles. The dataset contains over 5 million articles with silver multi-label annotated among 28 sectors by the data aggregator Factiva\footnote{\url{https://www.dowjones.com/professional/factiva/}}. Our collaborators have manually annotated 1,364 articles, of which 1,048 articles belonging to the 10 most frequent sectors are used for experiments in this paper. 

For these last two, let \x is the concatenation of the title, the summary, and the body text, and:
    \begin{align*}
        \T_0(\x) &= \text{Nouvelle \mask: $\x$}
        \\
        \T_1(\x) &= \text{Actualit\'e \mask: $\x$}
        \\
        \T_2(\x) &= \text{\mask: $\x$}
        \\
        \T_3(\x) &= \text{[Cat\'egorie: \mask] $\x$}
    \end{align*}

\subsection{Main Results}
\label{subsec:overall}

\begin{table*}[t]
    \centering
    \adjustbox{max width=\textwidth}{%
    \begin{tabular}{clcccccc}
        \hline
        $N$                         &{Verbalizer}         &\textbf{AG}                    &\textbf{DBpedia}                &\textbf{Yahoo}                &\textbf{\acrshort{frn}}        &\textbf{MLSUM Fr}           & {{Average}}\\
        \hline
        0                           &Majority
                                                        &\num{25.00}           &\num{07.14}            &\num{10.00}          &\num{16.79}           &\num{22.80}        & 16.36\\
                                    &Manual             &\num{72.14}           &\num{73.17}            &\bf\num{58.91}       &\bf\num{69.40}        &\num{51.45}        & \bf65.01\\
                                    &Soft               &\num{71.89}           &\num{54.57}            &\num{52.34}          &\num{64.74}           &\num{51.71}        & 59.05\\
                                    &MaVEN              &\num{72.75}           &\bf\num{74.77}         &\num{56.34}          &\num{62.69}           &\bf\num{54.52}     & 64.21        \\
                                    &Instruct           &\bf\num{75.58}        &\num{74.02}            &\num{52.42}          &\num{46.62}           &\num{37.47}        &\num{57.09}         \\           
        \hline
        32                          &Manual             &\num{83.96+-2.11}     &\num{91.68+-1.58}      &\num{61.84+-1.17}    &\num{81.16+-3.08}     &\num{58.42+-6.44}      & 75.41  \\
                                    &Soft               &\num{81.82+-3.30}     &\num{85.95+-1.12}      &\num{50.76+-2.84}    &\num{74.63+-5.54}     &\num{60.53+-4.86}      & 70.74\\
                                    &Auto               &\bf\num{86.44+-1.89}  &\num{79.24+-7.98}      &\num{50.08+-4.39}    &\num{73.63+-1.35}     &\num{56.38+-2.82}      & 69.15 \\
                                    &MaVEN              &\num{83.97+-2.70}     &\num{94.01+-1.08}   &\num{61.58+-3.46} &\bf\num{91.11+-1.68}     &\bf\num{60.81+-1.93}   &\bf78.30 \\
                                    &Instruct+ICL       &\num{82.25+-2.89}     &\bf\num{94.11+-0.97}   &\bf\num{63.58+-1.09} &\num{58.40+-2.16}     &\num{54.78+-1.83}      &70.62         \\
        \hline
        64                          &Manual             &\bf\num{88.14+-0.07}  &\num{96.75+-0.33}      &\num{65.29+-0.98}    &\num{90.17+-2.18}     &\num{65.79+-2.69}  & 81.23 \\
                                    &Soft               &\num{87.37+-0.45}     &\num{94.62+-2.06}      &\num{64.64+-1.10}    &\num{84.20+-0.88}     &\num{65.73+-2.68}  & 79.31\\
                                    &Auto               &\num{88.00+-0.46}     &\num{92.01+-2.92}      &\num{56.73+-5.05}    &\num{86.38+-3.64}     &\bf\num{67.17+-4.32} & 78.06 \\
                                    &MaVEN              &\num{87.57+-0.88}     &\bf\num{97.57+-0.29}   &\bf\num{66.17+-1.50} &\bf\num{90.49+-3.00}  &\num{65.88+-3.76}  & \bf81.54 \\
        \hline                  
        128                         &Manual             &\num{88.43+-0.33}     &\num{96.66+-1.14}      &\num{66.71+-0.61}    &\bf\num{94.28+-1.32}  &\num{69.13+-0.89}  & 83.04 \\
                                    &Soft               &\num{87.32+-0.56}     &\num{96.56+-2.00}      &\num{65.93+-0.86}    &\num{93.47+-2.44}     &\num{68.29+-0.84}  & 82.31 \\
                                    &Auto               &\bf\num{88.86+-0.10}  &\num{95.75+-1.87}      &\num{67.42+-0.36}    &\num{93.47+-0.56}     &\bf\num{71.28+-2.46}  & 83.36 \\
                                    &MaVEN              &\num{88.65+-0.57}     &\bf\num{97.85+-0.10}   &\bf\num{69.18+-0.66} &\num{93.28+-0.67}     &\num{68.22+-1.43}  & \bf83.44 \\
        \hline
    \end{tabular}
    }
    \caption{Accuracy of \acrshort{mave} compared to other verbalizers. The ensembling strategy is logit averaging. \textbf{Bold} are the best baselines. The last column is the average over five datasets. Our proposed \acrshort{mave} achieves significant performance gain compared to others for $N \in \{32, 64\}$ and best average performance for overall.}
    \label{tab:overall}
\end{table*}

\begin{figure*}[t]
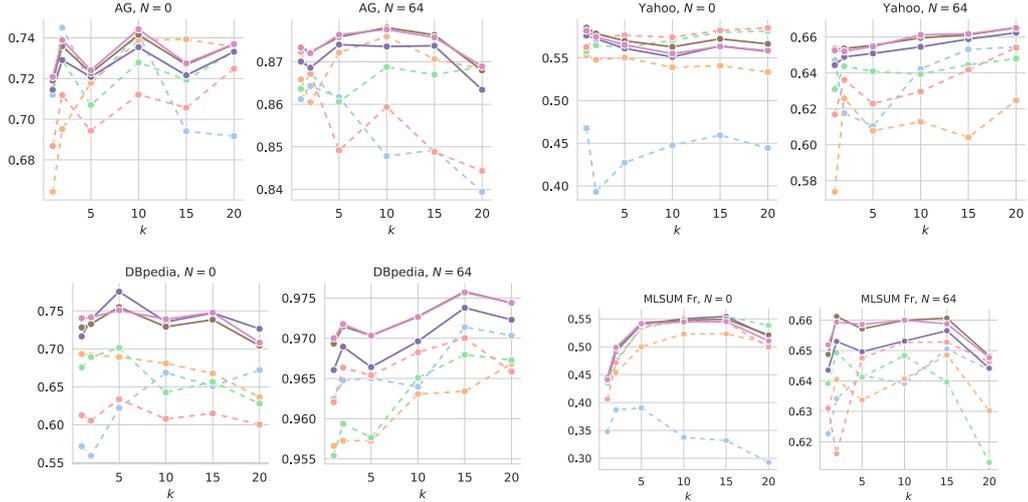

    \centering
    \includesvg[width=0.9\columnwidth]{figs/AG_k}
    \includesvg[width=0.9\columnwidth]{figs/Yahoo_k}
    \includesvg[width=0.9\columnwidth]{figs/DBpedia_k}
    \includesvg[width=0.8\columnwidth]{figs/MLSUM_Fr_k}
    \caption{Accuracy of \acrshort{mave} by number of label words, on four datasets for $N \in \{0, 64\}$. Dashed colored lines represent templates $T$ : \textcolor{C0}{0}, \textcolor{C1}{1}, \textcolor{C2}{2}, \textcolor{C3}{3}. Solid colored lines represent the ensemble methods:  \textcolor{C4}{vote}, \textcolor{C5}{proba}, \textcolor{C6}{logit}.}
    \label{fig:mave_k}
\end{figure*}

\Cref{tab:overall} shows the result over five datasets and three baselines, for different quantity of data $N$. 

For zero-shot learning, we observe that \acrshort{mave} achieves similar performance to the manual verbalizers, with the exception of \acrshort{frn}. We hypothesize that in this case, the neighborhoods of the class names do not model sufficiently the vocabulary of their classes without finetuning.


For extremely low-data settings, such as $N \in \{32,64\}$, we observe a clear superiority of \acrshort{mave}. Compared to the manual verbalizer, \acrshort{mave} achieves an improvement of $2.3$ on DBpedia, $10.0$ on FrN, and $2.4$ on MLSUM Fr for $N=32$. In other cases for $N\in \{32, 64\}$, \acrshort{mave} ranks as either the best or the second best among all verbalizers. For larger values of $N$, the gap between \acrshort{mave} manual verbalizer declines. Given more and more training data, the \acrshort{lm} learns to attribute the probability mass only to the core word, and thus, neighbor words become less useful. 


In summary, \acrshort{mave} consistently achieves the highest average score across five datasets all few-shot learning contexts. It shows an improvement of $2.9$ in average over the manual verbalizer for $N=32$. For the zero-shot case, it slightly underperforms the manual verbalizer. 

Additionally, we remark that for $N\ge 64$, the automatic verbalizer perform similarly, the manual verbalizer for all datasets (with $N \ge 32$ for AG and $N \ge 128$ for others). The main reason for this evolution is that the automatic algorithm mines for label words from likelihood on training data. With very few labeled data, the evaluation of this likelihood is less accurate. Notably, on AG and MLSUM Fr, the automatic verbalizer exceeds the manual verbalizer and \acrshort{mave}, which suggests that initial words given by the manual verbalizer of these datasets are biased and less accurate than words extracted from the data.

Compared to \acrshort{it} and \acrshort{icl} applied on \texttt{Mistral}, notice that combining \texttt{RoBERTa} or \texttt{Camembert} with verbalizers (\acrshort{mave} included) achieves a similar, sometimes higher, level of accuracy, despite having about 20 times fewer parameters (355M vs 7B). This finding encourages further research into optimizing smaller \acrshortpl{lm} to their fullest potential, rather than focusing on massively scaling the size and pretraining of \acrshortpl{llm}.

\subsection{Impact of the Neighborhood Size $k$}
\label{subsec:k}

Motivated by remarks in \cite{nguyen-etal-2024-enhancing} that using more label words produces stronger verbalizers, in this section, we inspect the impact of the parameter $k$ for our \acrshort{mave}.

\Cref{fig:mave_k} shows the prediction accuracy of individual models and assembled models with different $k$. For zero-shot prediction, the performance depends significantly on $k$, fluctuating within a range of $10.$ for MLSUM Fr and less than $5.$ for other datasets. With supervised data, fine-tuned models become more robust with $k$, where the variation is confined within a margin of about $2.$ globally, in particular around $0.6$ for DBpedia.

In practice, a fixed value between 10 and 15 guarantees a decent level of performance. We also observe that the dependence on $k$ is minor compared to the variation due to textual templates, discussed in \cref{subsec:ensemble}.

\subsection{Effectiveness of Ensemble Models}
\label{subsec:ensemble}


In \cref{fig:mave_k}, we assess outcomes by utilizing individual templates and by three different methods of ensemble. Generally, ensemble models yield more accurate predictions compared to using the most efficient template alone. Ensemble approaches not only improve prediction accuracy but also enhance stability and reduce the reliance on prompt selection, which typically relies on large validation sets \cite{perez2021true}, especially when individual templates show significant performance variations. Additionally, ensemble models are generally less sensitive to changes in the neighborhood size $k$, as discussed in \cref{subsec:k}.

Among the three methods, voting tends to be less effective than probability and logit averaging. However, this difference is minimal when compared to the overall improvement achieved by assembling individual templates.

\subsection{Effect of the Embedding Space $E$}

\begin{figure}[h]
    \centering
    \includesvg[width=0.9\columnwidth]{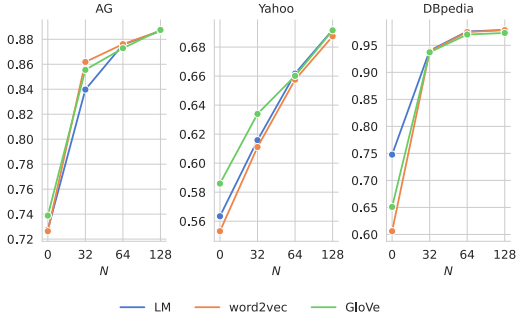}
    \caption{\acrshort{mave} accuracy using different embedding spaces (LM, word2vec, GloVe) with varying data amount $N$.}
    \label{fig:mave_E}
\end{figure}

In this section, we evaluate the influence of the embedding space $E$ to \acrshort{mave}. The embedding space intervenes in two manners: the choice of the neighborhood $\mathcal N_k(w_0)$ and the initialization of weights $q^y_w$ via $s(w_0, w)$ (\cref{sec:methodology}). The vanilla \acrshort{mave} utilizes the same embedding layer as the token embedding layer of the \acrshort{lm} (\texttt{RoBERTa-large} to be precise) as suggested by \cite{zhao-etal-2023-pre}, assuming the same embeddings as the fine-tuned \acrshort{lm} yields more coherence. To verify this tuition, \cref{fig:mave_E} demonstrates the performance of \acrshort{mave} using different embedding spaces: LM's embedding layer, Google word2vec\footnote{\url{https://code.google.com/archive/p/word2vec/}} \cite{mikolov2013distributed, mikolov2013efficient} and GloVe\footnote{\url{https://nlp.stanford.edu/projects/glove/}} pre-trained on Wikipedia and Gigaword \cite{pennington2014glove}.

In zero-shot, we observe a significant difference in performance. The range of variation is positively correlated to the number of classes for the considered problem. For example, the magnitude of this range of variation is approximately $1.$ for AG with 4 classes, $3.$ for Yahoo with 10 classes and up to $15.$ for DBpedia with 14 classes. Additionally, using the LM embedding surpasses word2vec and GloVe by a large margin on DBpedia, and works similarly to others in other cases.

When supervised data is available, we observe a convergent trend for the three embeddings. As the amount of data increases, the difference between models built from different embedding spaced reduces. For $N=128$, the score variation due to embedding space of \acrshort{mave} is less than $0.5$. The importance of the embedding space is minimized with the quantity of supervised data. 

An example of the neighborhood obtained from the different embeddings is in \cref{tab:E}, \cref{app:E}. For the LM embeddings, most extracted neighbors are spelling variants (e.g. ``Sport'' vs ``Sports''), case-sensitive variants (e.g. ``\_Sports'' vs ``\_sports'') or morphological variants (e.g. ``\_sports'' vs ``\_sport'') of the core tokens. In other cases, the neighborhood also includes tokens deriving from the same origin (e.g. ``science'', ``scientific'' and ``scientist''). This phenomenon is observed partly in GloVe and even less in word2vec. Tokens extracted from GloVe space are semantically related to the core tokens, providing global coverage of the topic of the considered class. Meanwhile, neighbors extracted by word2vec are rare combinations of words, proper nouns, etc., that are less meaningful. This could be a potential explanation for the poor performance of word2vec in \cref{fig:mave_E}.

\subsection{Sensitivity to the Initial Seed Label Words}

\label{sub:sensitivity}

\begin{figure}[t]
    \centering
    \includesvg[width=\columnwidth]{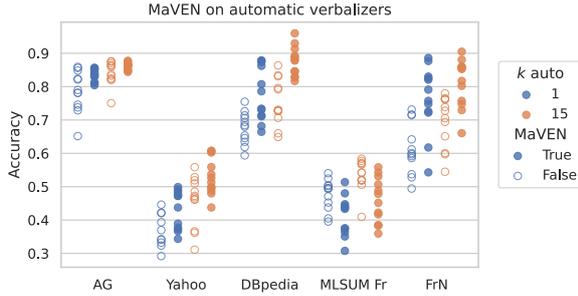}
    \caption{Accuracy of models initialized with automatic verbalizers, with and without \acrshort{mave}. Each point corresponds to one template under one random data split. All models are fine-tuned with $N=32$ examples.}
    \label{fig:AAR}
\end{figure}

As described in \cref{sec:methodology}, \acrshort{mave} relies on the manual label words used for initialization. The seed $w_0$ determines the neighborhood $\mathcal{N}_k(w_0)$, which in turn influences the selection of additional label words and their initial weights.

We propose a procedure to (i) find a reasonable initialization when manual seed words are not available and (ii) quantify the sensitivity of MaVEN's performance to varying initialization. First, we use the automatic verbalizer algorithm PETAL (\cref{sub:baselines}, \citealp{schick-etal-2020-automatically}) to extract $k_{\rm auto}$ label words for each class. The automatic verbalizer depends on the template and the training data, leading to different sets of core words for each run. This variation simulates the scenario of varying initial verbalizers that are relevant but not necessarily optimal for class representation. Next, these verbalizers are enriched using the \acrshort{mave} algorithm presented in \cref{sub:maven}. Finally, the augmented verbalizer and the LM are fine-tuned and evaluated as described in \cref{sec:experiment}. Comparing the augmented verbalizers with the initial verbalizers provides insights into the effectiveness of the proposed enrichment algorithm based on nearest neighbors.

\begin{figure}[t]
    \centering
    \includesvg[width=\columnwidth]{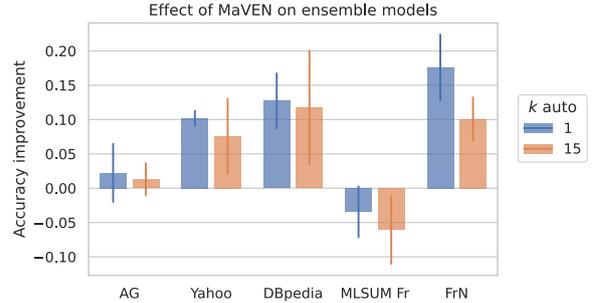}
    \caption{Improvement with \acrshort{mave} on logit-averaged models compared to their automatic initialization. All models are fine-tuned with $N=32$ examples.}
    \label{fig:delta}
\end{figure}

Experimental results in \cref{fig:AAR} for individual templates compare the performance of automatically initialized verbalizers with $k_{\rm auto}\in \{1, 15\}$, with and without MaVEN enrichment. \Cref{fig:delta} shows the improvement in accuracy upon applying \acrshort{mave}, evaluated on the ensemble models. We observe that MaVEN consistently contributes positively to the performance of automatic verbalizers on four out of five datasets. The exception for MLSUM Fr may be explained by the fact that the labels of this dataset is artificially created by topic grouping. The improvements of \acrshort{mave} is more visible for smaller $k_{\rm auto}$. Overall, the instability of the augmented verbalizers across templates and random seed is of the same order as that of the initial automatic vervalizers.

\section{Conclusion}

In this paper, we propose \acrshort{mave}, a novel method to extend the manual verbalizer that is effective for few-shot learning via prompt-based fine-tuning of \acrshortpl{plm}. By leveraging the neighborhood relationship in the embedding space of \acrshortpl{plm}, \acrshort{mave} was able to identify words related to the topic title to construct verbalizers without the need for data or external knowledge. Experiments show that \acrshort{mave} outperforms other constructions of verbalizer for extremely few-shot contexts.

\section{Discussion and Limitations}

As an extension of the manual verbalizer, \acrshort{mave} requires initial core words that contain the semantics meaning of the class. Therefore, theoretically, MaVEN is not applicable if class names are not meaningful descriptions of the classes. In reality, however, class titles often fully capture class concepts, and we rarely encounter cases where class titles are unavailable. The practicality of our proposed method remains. Otherwise, a substitute is proposed in \cref{sub:sensitivity}. In traditional fine-tuning where data amount is not limited, data instances represent classes. Meanwhile, in few-shot or zero-shot learning cases, class titles are the alternative representation of classes instead of data instances as in traditional fine-tuning.

The formulation and construction of verbalizers studied in this work focus on masked \acrshortpl{lm}, exploited only in encoder mode. Meanwhile, recent released \acrshortpl{plm} (GPT \citealp{brown2020language}, LLaMA \citealp{touvron2023llama}, Falcon \citealp{almazrouei2023falcon}, etc.) are auto-regressive models that are more powerful on a variety of benchmarks. This opens the potential to adapt verbalizer constructions for generative models in decode mode, to exploit the rich knowledge incorporated in these large \acrshortpl{lm}.

Our work includes datasets and verbalizers in English and French only. It is not guaranteed that the conclusions generalize well. Other works in other languages or more research on verbalizers with multi-lingual models can be explored.

\bibliography{custom,anthology}

\begin{thebibliography}{46}
\providecommand{\natexlab}[1]{#1}

\bibitem[{Almazrouei et~al.(2023)Almazrouei, Alobeidli, Alshamsi, Cappelli, Cojocaru, Debbah, Étienne Goffinet, Hesslow, Launay, Malartic, Mazzotta, Noune, Pannier, and Penedo}]{almazrouei2023falcon}
Ebtesam Almazrouei, Hamza Alobeidli, Abdulaziz Alshamsi, Alessandro Cappelli, Ruxandra Cojocaru, Mérouane Debbah, Étienne Goffinet, Daniel Hesslow, Julien Launay, Quentin Malartic, Daniele Mazzotta, Badreddine Noune, Baptiste Pannier, and Guilherme Penedo. 2023.
\newblock \href {https://arxiv.org/abs/2311.16867} {The falcon series of open language models}.
\newblock \emph{Preprint}, arXiv:2311.16867.

\bibitem[{Bach et~al.(2022)Bach, Sanh, Yong, Webson, Raffel, Nayak, Sharma, Kim, Bari, Fevry, Alyafeai, Dey, Santilli, Sun, Ben-David, Xu, Chhablani, Wang, Fries, Al-shaibani, Sharma, Thakker, Almubarak, Tang, Radev, Jiang, and Rush}]{bach2022promptsource}
Stephen~H. Bach, Victor Sanh, Zheng-Xin Yong, Albert Webson, Colin Raffel, Nihal~V. Nayak, Abheesht Sharma, Taewoon Kim, M~Saiful Bari, Thibault Fevry, Zaid Alyafeai, Manan Dey, Andrea Santilli, Zhiqing Sun, Srulik Ben-David, Canwen Xu, Gunjan Chhablani, Han Wang, Jason~Alan Fries, Maged~S. Al-shaibani, Shanya Sharma, Urmish Thakker, Khalid Almubarak, Xiangru Tang, Dragomir Radev, Mike Tian-Jian Jiang, and Alexander~M. Rush. 2022.
\newblock \href {https://arxiv.org/abs/2202.01279} {Promptsource: An integrated development environment and repository for natural language prompts}.
\newblock \emph{Preprint}, arXiv:2202.01279.

\bibitem[{Brown et~al.(2020)Brown, Mann, Ryder, Subbiah, Kaplan, Dhariwal, Neelakantan, Shyam, Sastry, Askell, Agarwal, Herbert-Voss, Krueger, Henighan, Child, Ramesh, Ziegler, Wu, Winter, Hesse, Chen, Sigler, Litwin, Gray, Chess, Clark, Berner, McCandlish, Radford, Sutskever, and Amodei}]{brown2020language}
Tom~B. Brown, Benjamin Mann, Nick Ryder, Melanie Subbiah, Jared Kaplan, Prafulla Dhariwal, Arvind Neelakantan, Pranav Shyam, Girish Sastry, Amanda Askell, Sandhini Agarwal, Ariel Herbert-Voss, Gretchen Krueger, Tom Henighan, Rewon Child, Aditya Ramesh, Daniel~M. Ziegler, Jeffrey Wu, Clemens Winter, Christopher Hesse, Mark Chen, Eric Sigler, Mateusz Litwin, Scott Gray, Benjamin Chess, Jack Clark, Christopher Berner, Sam McCandlish, Alec Radford, Ilya Sutskever, and Dario Amodei. 2020.
\newblock \href {https://arxiv.org/abs/2005.14165} {Language models are few-shot learners}.
\newblock \emph{Preprint}, arXiv:2005.14165.

\bibitem[{Chen et~al.(2022)Chen, Zhang, Xie, Deng, Yao, Tan, Huang, Si, and Chen}]{Chen_2022}
Xiang Chen, Ningyu Zhang, Xin Xie, Shumin Deng, Yunzhi Yao, Chuanqi Tan, Fei Huang, Luo Si, and Huajun Chen. 2022.
\newblock \href {https://doi.org/10.1145/3485447.3511998} {{KnowPrompt}: Knowledge-aware prompt-tuning with synergistic optimization for relation extraction}.
\newblock In \emph{Proceedings of the {ACM} Web Conference 2022}. {ACM}.

\bibitem[{Chung et~al.(2022)Chung, Hou, Longpre, Zoph, Tay, Fedus, Li, Wang, Dehghani, Brahma, Webson, Gu, Dai, Suzgun, Chen, Chowdhery, Castro-Ros, Pellat, Robinson, Valter, Narang, Mishra, Yu, Zhao, Huang, Dai, Yu, Petrov, Chi, Dean, Devlin, Roberts, Zhou, Le, and Wei}]{chung2022scaling}
Hyung~Won Chung, Le~Hou, Shayne Longpre, Barret Zoph, Yi~Tay, William Fedus, Yunxuan Li, Xuezhi Wang, Mostafa Dehghani, Siddhartha Brahma, Albert Webson, Shixiang~Shane Gu, Zhuyun Dai, Mirac Suzgun, Xinyun Chen, Aakanksha Chowdhery, Alex Castro-Ros, Marie Pellat, Kevin Robinson, Dasha Valter, Sharan Narang, Gaurav Mishra, Adams Yu, Vincent Zhao, Yanping Huang, Andrew Dai, Hongkun Yu, Slav Petrov, Ed~H. Chi, Jeff Dean, Jacob Devlin, Adam Roberts, Denny Zhou, Quoc~V. Le, and Jason Wei. 2022.
\newblock \href {https://arxiv.org/abs/2210.11416} {Scaling instruction-finetuned language models}.
\newblock \emph{Preprint}, arXiv:2210.11416.

\bibitem[{Cui et~al.(2022)Cui, Hu, Ding, Huang, and Liu}]{cui-etal-2022-prototypical}
Ganqu Cui, Shengding Hu, Ning Ding, Longtao Huang, and Zhiyuan Liu. 2022.
\newblock \href {https://doi.org/10.18653/v1/2022.acl-long.483} {Prototypical verbalizer for prompt-based few-shot tuning}.
\newblock In \emph{Proceedings of the 60th Annual Meeting of the Association for Computational Linguistics (Volume 1: Long Papers)}, pages 7014--7024, Dublin, Ireland. Association for Computational Linguistics.

\bibitem[{Devlin et~al.(2019{\natexlab{a}})Devlin, Chang, Lee, and Toutanova}]{devlin2019bert}
Jacob Devlin, Ming-Wei Chang, Kenton Lee, and Kristina Toutanova. 2019{\natexlab{a}}.
\newblock \href {https://arxiv.org/abs/1810.04805} {Bert: Pre-training of deep bidirectional transformers for language understanding}.
\newblock \emph{Preprint}, arXiv:1810.04805.

\bibitem[{Devlin et~al.(2019{\natexlab{b}})Devlin, Chang, Lee, and Toutanova}]{devlin-etal-2019-bert}
Jacob Devlin, Ming-Wei Chang, Kenton Lee, and Kristina Toutanova. 2019{\natexlab{b}}.
\newblock \href {https://doi.org/10.18653/v1/N19-1423} {{BERT}: Pre-training of deep bidirectional transformers for language understanding}.
\newblock In \emph{Proceedings of the 2019 Conference of the North {A}merican Chapter of the Association for Computational Linguistics: Human Language Technologies, Volume 1 (Long and Short Papers)}, pages 4171--4186, Minneapolis, Minnesota. Association for Computational Linguistics.

\bibitem[{Ding et~al.(2022)Ding, Hu, Zhao, Chen, Liu, Zheng, and Sun}]{ding-etal-2022-openprompt}
Ning Ding, Shengding Hu, Weilin Zhao, Yulin Chen, Zhiyuan Liu, Haitao Zheng, and Maosong Sun. 2022.
\newblock \href {https://doi.org/10.18653/v1/2022.acl-demo.10} {{O}pen{P}rompt: An open-source framework for prompt-learning}.
\newblock In \emph{Proceedings of the 60th Annual Meeting of the Association for Computational Linguistics: System Demonstrations}, pages 105--113, Dublin, Ireland. Association for Computational Linguistics.

\bibitem[{Dong et~al.(2023)Dong, Li, Dai, Zheng, Wu, Chang, Sun, Xu, Li, and Sui}]{dong2023survey}
Qingxiu Dong, Lei Li, Damai Dai, Ce~Zheng, Zhiyong Wu, Baobao Chang, Xu~Sun, Jingjing Xu, Lei Li, and Zhifang Sui. 2023.
\newblock \href {https://arxiv.org/abs/2301.00234} {A survey on in-context learning}.
\newblock \emph{Preprint}, arXiv:2301.00234.

\bibitem[{Gao et~al.(2021)Gao, Fisch, and Chen}]{gao-etal-2021-making}
Tianyu Gao, Adam Fisch, and Danqi Chen. 2021.
\newblock \href {https://doi.org/10.18653/v1/2021.acl-long.295} {Making pre-trained language models better few-shot learners}.
\newblock In \emph{Proceedings of the 59th Annual Meeting of the Association for Computational Linguistics and the 11th International Joint Conference on Natural Language Processing (Volume 1: Long Papers)}, pages 3816--3830, Online. Association for Computational Linguistics.

\bibitem[{Hambardzumyan et~al.(2021)Hambardzumyan, Khachatrian, and May}]{hambardzumyan-etal-2021-warp}
Karen Hambardzumyan, Hrant Khachatrian, and Jonathan May. 2021.
\newblock \href {https://doi.org/10.18653/v1/2021.acl-long.381} {{WARP}: {W}ord-level {A}dversarial {R}e{P}rogramming}.
\newblock In \emph{Proceedings of the 59th Annual Meeting of the Association for Computational Linguistics and the 11th International Joint Conference on Natural Language Processing (Volume 1: Long Papers)}, pages 4921--4933, Online. Association for Computational Linguistics.

\bibitem[{Hu et~al.(2022)Hu, Ding, Wang, Liu, Wang, Li, Wu, and Sun}]{hu-etal-2022-knowledgeable}
Shengding Hu, Ning Ding, Huadong Wang, Zhiyuan Liu, Jingang Wang, Juanzi Li, Wei Wu, and Maosong Sun. 2022.
\newblock \href {https://doi.org/10.18653/v1/2022.acl-long.158} {Knowledgeable prompt-tuning: Incorporating knowledge into prompt verbalizer for text classification}.
\newblock In \emph{Proceedings of the 60th Annual Meeting of the Association for Computational Linguistics (Volume 1: Long Papers)}, pages 2225--2240, Dublin, Ireland. Association for Computational Linguistics.

\bibitem[{Jiang et~al.(2023)Jiang, Sablayrolles, Mensch, Bamford, Chaplot, de~las Casas, Bressand, Lengyel, Lample, Saulnier, Lavaud, Lachaux, Stock, Scao, Lavril, Wang, Lacroix, and Sayed}]{jiang2023mistral}
Albert~Q. Jiang, Alexandre Sablayrolles, Arthur Mensch, Chris Bamford, Devendra~Singh Chaplot, Diego de~las Casas, Florian Bressand, Gianna Lengyel, Guillaume Lample, Lucile Saulnier, Lélio~Renard Lavaud, Marie-Anne Lachaux, Pierre Stock, Teven~Le Scao, Thibaut Lavril, Thomas Wang, Timothée Lacroix, and William~El Sayed. 2023.
\newblock \href {https://arxiv.org/abs/2310.06825} {Mistral 7b}.
\newblock \emph{Preprint}, arXiv:2310.06825.

\bibitem[{Jiang et~al.(2020)Jiang, Xu, Araki, and Neubig}]{jiang-etal-2020-know}
Zhengbao Jiang, Frank~F. Xu, Jun Araki, and Graham Neubig. 2020.
\newblock \href {https://doi.org/10.1162/tacl_a_00324} {How can we know what language models know?}
\newblock \emph{Transactions of the Association for Computational Linguistics}, 8:423--438.

\bibitem[{Lester et~al.(2021)Lester, Al-Rfou, and Constant}]{lester-etal-2021-power}
Brian Lester, Rami Al-Rfou, and Noah Constant. 2021.
\newblock \href {https://doi.org/10.18653/v1/2021.emnlp-main.243} {The power of scale for parameter-efficient prompt tuning}.
\newblock In \emph{Proceedings of the 2021 Conference on Empirical Methods in Natural Language Processing}, pages 3045--3059, Online and Punta Cana, Dominican Republic. Association for Computational Linguistics.

\bibitem[{Li and Liang(2021)}]{li-liang-2021-prefix}
Xiang~Lisa Li and Percy Liang. 2021.
\newblock \href {https://doi.org/10.18653/v1/2021.acl-long.353} {Prefix-tuning: Optimizing continuous prompts for generation}.
\newblock In \emph{Proceedings of the 59th Annual Meeting of the Association for Computational Linguistics and the 11th International Joint Conference on Natural Language Processing (Volume 1: Long Papers)}, pages 4582--4597, Online. Association for Computational Linguistics.

\bibitem[{Liu et~al.(2022)Liu, Ji, Fu, Tam, Du, Yang, and Tang}]{liu-etal-2022-p}
Xiao Liu, Kaixuan Ji, Yicheng Fu, Weng Tam, Zhengxiao Du, Zhilin Yang, and Jie Tang. 2022.
\newblock \href {https://doi.org/10.18653/v1/2022.acl-short.8} {{P}-tuning: Prompt tuning can be comparable to fine-tuning across scales and tasks}.
\newblock In \emph{Proceedings of the 60th Annual Meeting of the Association for Computational Linguistics (Volume 2: Short Papers)}, pages 61--68, Dublin, Ireland. Association for Computational Linguistics.

\bibitem[{Liu et~al.(2019)Liu, Ott, Goyal, Du, Joshi, Chen, Levy, Lewis, Zettlemoyer, and Stoyanov}]{liu2019roberta}
Yinhan Liu, Myle Ott, Naman Goyal, Jingfei Du, Mandar Joshi, Danqi Chen, Omer Levy, Mike Lewis, Luke Zettlemoyer, and Veselin Stoyanov. 2019.
\newblock \href {https://arxiv.org/abs/1907.11692} {Roberta: A robustly optimized bert pretraining approach}.
\newblock \emph{Preprint}, arXiv:1907.11692.

\bibitem[{Martin et~al.(2020)Martin, Muller, Ortiz~Su{\'a}rez, Dupont, Romary, de~la Clergerie, Seddah, and Sagot}]{martin-etal-2020-camembert}
Louis Martin, Benjamin Muller, Pedro~Javier Ortiz~Su{\'a}rez, Yoann Dupont, Laurent Romary, {\'E}ric de~la Clergerie, Djam{\'e} Seddah, and Beno{\^\i}t Sagot. 2020.
\newblock \href {https://doi.org/10.18653/v1/2020.acl-main.645} {{C}amem{BERT}: a tasty {F}rench language model}.
\newblock In \emph{Proceedings of the 58th Annual Meeting of the Association for Computational Linguistics}, pages 7203--7219, Online. Association for Computational Linguistics.

\bibitem[{Mikolov et~al.(2013{\natexlab{a}})Mikolov, Chen, Corrado, and Dean}]{mikolov2013efficient}
Tomas Mikolov, Kai Chen, Greg Corrado, and Jeffrey Dean. 2013{\natexlab{a}}.
\newblock \href {https://arxiv.org/abs/1301.3781} {Efficient estimation of word representations in vector space}.
\newblock \emph{Preprint}, arXiv:1301.3781.

\bibitem[{Mikolov et~al.(2013{\natexlab{b}})Mikolov, Sutskever, Chen, Corrado, and Dean}]{mikolov2013distributed}
Tomas Mikolov, Ilya Sutskever, Kai Chen, Greg Corrado, and Jeffrey Dean. 2013{\natexlab{b}}.
\newblock \href {https://arxiv.org/abs/1310.4546} {Distributed representations of words and phrases and their compositionality}.
\newblock \emph{Preprint}, arXiv:1310.4546.

\bibitem[{Miller(1994)}]{miller-1994-wordnet}
George~A. Miller. 1994.
\newblock \href {https://aclanthology.org/H94-1111} {{W}ord{N}et: A lexical database for {E}nglish}.
\newblock In \emph{{H}uman {L}anguage {T}echnology: Proceedings of a Workshop held at {P}lainsboro, {N}ew {J}ersey, {M}arch 8-11, 1994}.

\bibitem[{Muennighoff et~al.(2023)Muennighoff, Wang, Sutawika, Roberts, Biderman, Scao, Bari, Shen, Yong, Schoelkopf, Tang, Radev, Aji, Almubarak, Albanie, Alyafeai, Webson, Raff, and Raffel}]{muennighoff2023crosslingual}
Niklas Muennighoff, Thomas Wang, Lintang Sutawika, Adam Roberts, Stella Biderman, Teven~Le Scao, M~Saiful Bari, Sheng Shen, Zheng-Xin Yong, Hailey Schoelkopf, Xiangru Tang, Dragomir Radev, Alham~Fikri Aji, Khalid Almubarak, Samuel Albanie, Zaid Alyafeai, Albert Webson, Edward Raff, and Colin Raffel. 2023.
\newblock \href {https://arxiv.org/abs/2211.01786} {Crosslingual generalization through multitask finetuning}.
\newblock \emph{Preprint}, arXiv:2211.01786.

\bibitem[{Nguyen et~al.(2024)Nguyen, Tomeh, Lebbah, Charnois, Azzag, and Cordoba~Mu{\~n}oz}]{nguyen-etal-2024-enhancing}
Quang~Anh Nguyen, Nadi Tomeh, Mustapha Lebbah, Thierry Charnois, Hanene Azzag, and Santiago Cordoba~Mu{\~n}oz. 2024.
\newblock \href {https://aclanthology.org/2024.lrec-main.527} {Enhancing few-shot topic classification with verbalizers. a study on automatic verbalizer and ensemble methods}.
\newblock In \emph{Proceedings of the 2024 Joint International Conference on Computational Linguistics, Language Resources and Evaluation (LREC-COLING 2024)}, pages 5956--5965, Torino, Italia. ELRA and ICCL.

\bibitem[{Ouyang et~al.(2022)Ouyang, Wu, Jiang, Almeida, Wainwright, Mishkin, Zhang, Agarwal, Slama, Ray, Schulman, Hilton, Kelton, Miller, Simens, Askell, Welinder, Christiano, Leike, and Lowe}]{ouyang2022training}
Long Ouyang, Jeff Wu, Xu~Jiang, Diogo Almeida, Carroll~L. Wainwright, Pamela Mishkin, Chong Zhang, Sandhini Agarwal, Katarina Slama, Alex Ray, John Schulman, Jacob Hilton, Fraser Kelton, Luke Miller, Maddie Simens, Amanda Askell, Peter Welinder, Paul Christiano, Jan Leike, and Ryan Lowe. 2022.
\newblock \href {https://arxiv.org/abs/2203.02155} {Training language models to follow instructions with human feedback}.
\newblock \emph{Preprint}, arXiv:2203.02155.

\bibitem[{Pennington et~al.(2014)Pennington, Socher, and Manning}]{pennington2014glove}
Jeffrey Pennington, Richard Socher, and Christopher~D. Manning. 2014.
\newblock \href {http://www.aclweb.org/anthology/D14-1162} {Glove: Global vectors for word representation}.
\newblock In \emph{Empirical Methods in Natural Language Processing (EMNLP)}, pages 1532--1543.

\bibitem[{Perez et~al.(2021)Perez, Kiela, and Cho}]{perez2021true}
Ethan Perez, Douwe Kiela, and Kyunghyun Cho. 2021.
\newblock \href {https://arxiv.org/abs/2105.11447} {True few-shot learning with language models}.
\newblock \emph{Preprint}, arXiv:2105.11447.

\bibitem[{Petroni et~al.(2019)Petroni, Rockt{\"a}schel, Riedel, Lewis, Bakhtin, Wu, and Miller}]{petroni-etal-2019-language}
Fabio Petroni, Tim Rockt{\"a}schel, Sebastian Riedel, Patrick Lewis, Anton Bakhtin, Yuxiang Wu, and Alexander Miller. 2019.
\newblock \href {https://doi.org/10.18653/v1/D19-1250} {Language models as knowledge bases?}
\newblock In \emph{Proceedings of the 2019 Conference on Empirical Methods in Natural Language Processing and the 9th International Joint Conference on Natural Language Processing (EMNLP-IJCNLP)}, pages 2463--2473, Hong Kong, China. Association for Computational Linguistics.

\bibitem[{Raffel et~al.(2020)Raffel, Shazeer, Roberts, Lee, Narang, Matena, Zhou, Li, and Liu}]{JMLR:v21:20-074}
Colin Raffel, Noam Shazeer, Adam Roberts, Katherine Lee, Sharan Narang, Michael Matena, Yanqi Zhou, Wei Li, and Peter~J. Liu. 2020.
\newblock \href {http://jmlr.org/papers/v21/20-074.html} {Exploring the limits of transfer learning with a unified text-to-text transformer}.
\newblock \emph{Journal of Machine Learning Research}, 21(140):1--67.

\bibitem[{Sanh et~al.(2022)Sanh, Webson, Raffel, Bach, Sutawika, Alyafeai, Chaffin, Stiegler, Scao, Raja, Dey, Bari, Xu, Thakker, Sharma, Szczechla, Kim, Chhablani, Nayak, Datta, Chang, Jiang, Wang, Manica, Shen, Yong, Pandey, Bawden, Wang, Neeraj, Rozen, Sharma, Santilli, Fevry, Fries, Teehan, Bers, Biderman, Gao, Wolf, and Rush}]{sanh2022multitask}
Victor Sanh, Albert Webson, Colin Raffel, Stephen~H. Bach, Lintang Sutawika, Zaid Alyafeai, Antoine Chaffin, Arnaud Stiegler, Teven~Le Scao, Arun Raja, Manan Dey, M~Saiful Bari, Canwen Xu, Urmish Thakker, Shanya~Sharma Sharma, Eliza Szczechla, Taewoon Kim, Gunjan Chhablani, Nihal Nayak, Debajyoti Datta, Jonathan Chang, Mike Tian-Jian Jiang, Han Wang, Matteo Manica, Sheng Shen, Zheng~Xin Yong, Harshit Pandey, Rachel Bawden, Thomas Wang, Trishala Neeraj, Jos Rozen, Abheesht Sharma, Andrea Santilli, Thibault Fevry, Jason~Alan Fries, Ryan Teehan, Tali Bers, Stella Biderman, Leo Gao, Thomas Wolf, and Alexander~M. Rush. 2022.
\newblock \href {https://arxiv.org/abs/2110.08207} {Multitask prompted training enables zero-shot task generalization}.
\newblock \emph{Preprint}, arXiv:2110.08207.

\bibitem[{Schick et~al.(2020)Schick, Schmid, and Sch{\"u}tze}]{schick-etal-2020-automatically}
Timo Schick, Helmut Schmid, and Hinrich Sch{\"u}tze. 2020.
\newblock \href {https://doi.org/10.18653/v1/2020.coling-main.488} {Automatically identifying words that can serve as labels for few-shot text classification}.
\newblock In \emph{Proceedings of the 28th International Conference on Computational Linguistics}, pages 5569--5578, Barcelona, Spain (Online). International Committee on Computational Linguistics.

\bibitem[{Schick and Sch{\"u}tze(2021{\natexlab{a}})}]{schick-schutze-2021-exploiting}
Timo Schick and Hinrich Sch{\"u}tze. 2021{\natexlab{a}}.
\newblock \href {https://doi.org/10.18653/v1/2021.eacl-main.20} {Exploiting cloze-questions for few-shot text classification and natural language inference}.
\newblock In \emph{Proceedings of the 16th Conference of the European Chapter of the Association for Computational Linguistics: Main Volume}, pages 255--269, Online. Association for Computational Linguistics.

\bibitem[{Schick and Sch{\"u}tze(2021{\natexlab{b}})}]{schick-schutze-2021-just}
Timo Schick and Hinrich Sch{\"u}tze. 2021{\natexlab{b}}.
\newblock \href {https://doi.org/10.18653/v1/2021.naacl-main.185} {It{'}s not just size that matters: Small language models are also few-shot learners}.
\newblock In \emph{Proceedings of the 2021 Conference of the North American Chapter of the Association for Computational Linguistics: Human Language Technologies}, pages 2339--2352, Online. Association for Computational Linguistics.

\bibitem[{Scialom et~al.(2020)Scialom, Dray, Lamprier, Piwowarski, and Staiano}]{scialom-etal-2020-mlsum}
Thomas Scialom, Paul-Alexis Dray, Sylvain Lamprier, Benjamin Piwowarski, and Jacopo Staiano. 2020.
\newblock \href {https://doi.org/10.18653/v1/2020.emnlp-main.647} {{MLSUM}: The multilingual summarization corpus}.
\newblock In \emph{Proceedings of the 2020 Conference on Empirical Methods in Natural Language Processing (EMNLP)}, pages 8051--8067, Online. Association for Computational Linguistics.

\bibitem[{Speer and Havasi(2012)}]{speer-havasi-2012-representing}
Robyn Speer and Catherine Havasi. 2012.
\newblock \href {http://www.lrec-conf.org/proceedings/lrec2012/pdf/1072_Paper.pdf} {Representing general relational knowledge in {C}oncept{N}et 5}.
\newblock In \emph{Proceedings of the Eighth International Conference on Language Resources and Evaluation ({LREC}'12)}, pages 3679--3686, Istanbul, Turkey. European Language Resources Association (ELRA).

\bibitem[{Touvron et~al.(2023)Touvron, Lavril, Izacard, Martinet, Lachaux, Lacroix, Rozière, Goyal, Hambro, Azhar, Rodriguez, Joulin, Grave, and Lample}]{touvron2023llama}
Hugo Touvron, Thibaut Lavril, Gautier Izacard, Xavier Martinet, Marie-Anne Lachaux, Timothée Lacroix, Baptiste Rozière, Naman Goyal, Eric Hambro, Faisal Azhar, Aurelien Rodriguez, Armand Joulin, Edouard Grave, and Guillaume Lample. 2023.
\newblock \href {https://arxiv.org/abs/2302.13971} {Llama: Open and efficient foundation language models}.
\newblock \emph{Preprint}, arXiv:2302.13971.

\bibitem[{Wei et~al.(2022)Wei, Bosma, Zhao, Guu, Yu, Lester, Du, Dai, and Le}]{wei2022finetuned}
Jason Wei, Maarten Bosma, Vincent~Y. Zhao, Kelvin Guu, Adams~Wei Yu, Brian Lester, Nan Du, Andrew~M. Dai, and Quoc~V. Le. 2022.
\newblock \href {https://arxiv.org/abs/2109.01652} {Finetuned language models are zero-shot learners}.
\newblock \emph{Preprint}, arXiv:2109.01652.

\bibitem[{Wolf et~al.(2020)Wolf, Debut, Sanh, Chaumond, Delangue, Moi, Cistac, Rault, Louf, Funtowicz, Davison, Shleifer, von Platen, Ma, Jernite, Plu, Xu, Le~Scao, Gugger, Drame, Lhoest, and Rush}]{wolf-etal-2020-transformers}
Thomas Wolf, Lysandre Debut, Victor Sanh, Julien Chaumond, Clement Delangue, Anthony Moi, Pierric Cistac, Tim Rault, Remi Louf, Morgan Funtowicz, Joe Davison, Sam Shleifer, Patrick von Platen, Clara Ma, Yacine Jernite, Julien Plu, Canwen Xu, Teven Le~Scao, Sylvain Gugger, Mariama Drame, Quentin Lhoest, and Alexander Rush. 2020.
\newblock \href {https://doi.org/10.18653/v1/2020.emnlp-demos.6} {Transformers: State-of-the-art natural language processing}.
\newblock In \emph{Proceedings of the 2020 Conference on Empirical Methods in Natural Language Processing: System Demonstrations}, pages 38--45, Online. Association for Computational Linguistics.

\bibitem[{Yin et~al.(2019)Yin, Hay, and Roth}]{yin-etal-2019-benchmarking}
Wenpeng Yin, Jamaal Hay, and Dan Roth. 2019.
\newblock \href {https://doi.org/10.18653/v1/D19-1404} {Benchmarking zero-shot text classification: Datasets, evaluation and entailment approach}.
\newblock In \emph{Proceedings of the 2019 Conference on Empirical Methods in Natural Language Processing and the 9th International Joint Conference on Natural Language Processing (EMNLP-IJCNLP)}, pages 3914--3923, Hong Kong, China. Association for Computational Linguistics.

\bibitem[{Zhang et~al.(2023)Zhang, Liang, Zhan, Wu, and Lam}]{zhang-etal-2023-revisit}
Haode Zhang, Haowen Liang, Li-Ming Zhan, Xiao-Ming Wu, and Albert~Y.S. Lam. 2023.
\newblock \href {https://doi.org/10.18653/v1/2023.findings-acl.706} {Revisit few-shot intent classification with {PLM}s: Direct fine-tuning vs. continual pre-training}.
\newblock In \emph{Findings of the Association for Computational Linguistics: ACL 2023}, pages 11105--11121, Toronto, Canada. Association for Computational Linguistics.

\bibitem[{Zhang et~al.(2024)Zhang, Dong, Li, Zhang, Sun, Wang, Li, Hu, Zhang, Wu, and Wang}]{zhang2024instruction}
Shengyu Zhang, Linfeng Dong, Xiaoya Li, Sen Zhang, Xiaofei Sun, Shuhe Wang, Jiwei Li, Runyi Hu, Tianwei Zhang, Fei Wu, and Guoyin Wang. 2024.
\newblock \href {https://arxiv.org/abs/2308.10792} {Instruction tuning for large language models: A survey}.
\newblock \emph{Preprint}, arXiv:2308.10792.

\bibitem[{Zhang et~al.(2015)Zhang, Zhao, and LeCun}]{NIPS2015_250cf8b5}
Xiang Zhang, Junbo Zhao, and Yann LeCun. 2015.
\newblock \href {https://proceedings.neurips.cc/paper_files/paper/2015/file/250cf8b51c773f3f8dc8b4be867a9a02-Paper.pdf} {Character-level convolutional networks for text classification}.
\newblock In \emph{Advances in Neural Information Processing Systems}, volume~28. Curran Associates, Inc.

\bibitem[{Zhao et~al.(2023)Zhao, Ouyang, Yu, Wu, and Li}]{zhao-etal-2023-pre}
Xuandong Zhao, Siqi Ouyang, Zhiguo Yu, Ming Wu, and Lei Li. 2023.
\newblock \href {https://doi.org/10.18653/v1/2023.acl-long.869} {Pre-trained language models can be fully zero-shot learners}.
\newblock In \emph{Proceedings of the 61st Annual Meeting of the Association for Computational Linguistics (Volume 1: Long Papers)}, pages 15590--15606, Toronto, Canada. Association for Computational Linguistics.

\bibitem[{Zheng et~al.(2022)Zheng, Zhou, Qian, Ding, Liao, Jian, Salakhutdinov, Tang, Ruder, and Yang}]{zheng-etal-2022-fewnlu}
Yanan Zheng, Jing Zhou, Yujie Qian, Ming Ding, Chonghua Liao, Li~Jian, Ruslan Salakhutdinov, Jie Tang, Sebastian Ruder, and Zhilin Yang. 2022.
\newblock \href {https://doi.org/10.18653/v1/2022.acl-long.38} {{F}ew{NLU}: Benchmarking state-of-the-art methods for few-shot natural language understanding}.
\newblock In \emph{Proceedings of the 60th Annual Meeting of the Association for Computational Linguistics (Volume 1: Long Papers)}, pages 501--516, Dublin, Ireland. Association for Computational Linguistics.

\bibitem[{Zhuang et~al.(2021)Zhuang, Wayne, Ya, and Jun}]{zhuang-etal-2021-robustly}
Liu Zhuang, Lin Wayne, Shi Ya, and Zhao Jun. 2021.
\newblock \href {https://aclanthology.org/2021.ccl-1.108} {A robustly optimized {BERT} pre-training approach with post-training}.
\newblock In \emph{Proceedings of the 20th Chinese National Conference on Computational Linguistics}, pages 1218--1227, Huhhot, China. Chinese Information Processing Society of China.

\end{thebibliography}

\appendix

\section{Hyperparameters}
For simplicity, most choices of hyperparameters are based on existing works and practical considerations. However, these choices could have been done using the validation set.
\label{app:params}
\begin{table}[h]
    \centering
    \begin{tabular}{lr}
    \hline
        Parameter                       & Value                     \\
        \hline
        Optimizer                       & AdamW                     \\
        Learning rate\footnote{The learning rate increases linearly from 0 to its maximal value for the first 10\% steps, then decreases linearly to 0.}
        & $\num{1e-5}$              \\
        Training epochs                 & 10                        \\
        Batch size                      & 4                         \\
        Weight decay                    & 0.01                      \\
        $\beta_1$                       & 0.9                       \\
        $\beta_2$                       & 0.999                     \\
        Gradient accumulation           & 1                         \\
        
    \hline
    \end{tabular}
    \caption{Hyperparameters for fine-tuning.}
    \label{tab:params}
\end{table}

\section{Manual Verbalizers}
\label{app:mv}

Here, we specify the label words used for the manual verbalizers of each dataset in \cref{tab:mven} and \cref{tab:mvfr}.

\begin{table*}[p]
\centering
\begin{tabular}{ll}
\hline
Dataset \& Classes                            & Label words                       \\
\hline
\bf AG\\
        World                             & world, politics                    \\
        Sports                            & sports                             \\
        Business                          & business                           \\
        Sci/Tech                          & science, technology                \\
\hline
\bf DBpedia\\
        Company                           & company                            \\
        EducationalInstitution            & educational, institution           \\
        Artist                            & artist                             \\
        Athlete                           & athlete, sport                     \\
        OfficeHolder                      & office                             \\
        MeanOfTransportation              & transportaion                      \\
        Building                          & building                           \\
        NaturalPlace                      & natural, place                     \\
        Village                           & village                            \\
        Animal                            & animal                             \\
        Plant                             & plant                              \\
        Album                             & album                              \\
        Film                              & film                               \\
        WrittenWork                       & written, work                      \\
\hline
\bf Yahoo\\
        Society \& Culture                & society, culture,                  \\
        Science \& Mathematics            & science, mathematics               \\
        Health                            & health                             \\
        Education \& Reference            & education, reference               \\
        Computers \& Internet             & computers, internet                \\
        Sports                            & sports                             \\
        Business \& Finance               & business, finance                  \\
        Entertainment \& Music            & entertainment, music               \\
        Family \& Relationships           & family, relationships              \\
        Politics \& Government            & politics, government               \\
\hline
\bf MLSUM Fr\\
        Economie                          & économie                           \\
        Opinion                           & opinion                            \\
        Politique                         & politique                          \\
        Societe                           & société                            \\
        Culture                           & culture                            \\
        Sport                             & sport                              \\
        Environnement                     & environnement                       \\
        Technologie                       & technologie                        \\
        Education                         & éducation                          \\
        Justice                           & justice                            \\
\hline
\end{tabular}
\caption{Manual verbalizers of AG, DBPedia, Yahoo, and MLSUM Fr.}
\label{tab:mven}
\end{table*}

\begin{table*}[p]
\centering
\begin{tabularx}{\textwidth}{p{0.2\textwidth}X}
\hline
Class                            & Label words                        \\
\hline
        AERONAUTIQUE-ARMEMENT             & \textbf{aéronautique}, \textbf{armement}, flotte, rafale, marine, spatiale, pilote, défense, fusil, satellites, combat, missiles, militaire, réacteurs, hypersonique              \\
        AGRO-ALIMENTAIRE                  & \textbf{agroalimentaire}, \textbf{agriculture}, agricole, FAO, viticulture, sécheresse, plantation, biodiversité, {alimentation}, rurale, récolte, bio, terroir, paysanne, céréaliers           \\
        AUTOMOBILE                        & \textbf{automobile}, auto, carrosserie, voiture, motorisation, conduite, diesel, pney, mécanique, mobilité, Volkswagen, Renault, berline, concessions, SUV                         \\
        DISTRIBUTION-COMMERCE             & \textbf{distribution}, \textbf{commerce}, boutique, retail, vitrine, caisse, e-commerce, hypermarchés, ventes, distributeur, soldes, magasin, supermarchés, commercial, dropshipping            \\
        ELECTRICITE                       & \textbf{électricité}, energie, energy, éolienne, energetique, photovoltaique, nucléaire, gaz, carbone, combustion, solaire, électronique, generation, centrailes, hydrogène                       \\
        FINANCE                           & \textbf{finance}, banque, bancaire, monétaire, bce, solvabilité, liquidité, bale, financière, dette, holding, investisseur, investissement, capital, prêts                            \\
        PETROLE-GAZ                       & \textbf{pétrole}, \textbf{gaz}, energie, pétrolière, combustion, géo, forage, réserves, pipeline, oléoduc, gazoduc, rafinerie, liquefié, gisement, bitumeux                       \\
        PIM                               & PIM, \textbf{immobilier}, foncière, gestion, biens, proprieté, location, \textbf{promotion}, projets, permis, programmes, promoteurs, immeubles, chantiers, aménageurs             \\
        TOURISME-HOTELLERIE-RESTAURATION  & \textbf{tourisme}, \textbf{hôtellerie}, \textbf{restauration}, hotel, restaurant, vacances, vacanciers, séjour, auberges, camping, attraction, touristique, parc, croisiéristes, réservations\\
        TRANSPORT                         & \textbf{transport}, avion, bateaux, ferroviaire, douane, circulation, passagers, aérien, terrestre, maritime, conteneurs, navires, cargos, aéroport, fret                        \\
\hline
\end{tabularx}
\caption{Manual verbalizer of \acrshort{frn}, provided by our private company collaborator. \textbf{Bold} words indicates in \texttt{title} \cref{fig:preliminary}.}
\label{tab:mvfr}
\end{table*}

\section{Preliminary Experiments on \acrshort{frn}}
\label{app:pre}
\begin{figure}[h]
    \centering
    \includesvg[width=\columnwidth]{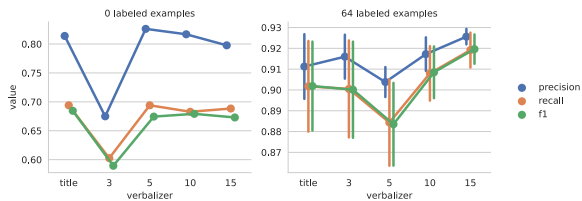}
    \caption{Study of different sizes for the manual verbalizer on the \acrshort{frn} dataset. \texttt{title} means using words in class names as label words.}
    \label{fig:preliminary}
\end{figure}

We examine the \acrshort{frn} dataset in zero-shot and in few-shot context with $N=64$, with the manual verbalizer provided by our collaborators of 15 words per class. By retaining the $k$ most important words (see \cref{tab:mvfr}), we observe the influence of the number of label words. \Cref{fig:preliminary} shows a clear improvement from 5 label words for zero-shot and 10 for few-shot. Moreover, few-shot models are more stable with more label words. This correlation is highly dependent on the ordering of importance of $v(y)$, therefore on human decision. However, the observation motivates us to inspect this phenomenon for an automatic search algorithm, such as PETAL or \acrshort{mave}.

\section{Examples of Neighborhood with Different Embeddings}
\label{app:E}

\Cref{tab:E} presents the neighborhood of 15 nearest tokens provided by three embedding spaces for two example core words ``sports'' and ``science''. 

\begin{table*}[t]
\footnotesize
\ttfamily
\begin{tabularx}{\textwidth}{X|lp{1cm}|lp{1cm}|lp{1cm}}
\hline
\rmfamily Embedding
        & \rmfamily LM             
        &        & \rmfamily word2vec                                                   &          & \rmfamily GloVe        &        \\
\hline
sports  & \_Sports         & 0.7727 & sport                                                      & 0.6915   & sport     & 0.7274\\
        & \_sport          & 0.7537 & sporting                                                   & 0.6360   & sporting      & 0.5801\\
        & \_sporting       & 0.6824 & Sports                                                     & 0.6295   & basketball   & 0.5788\\
        & \_athletics      & 0.6536 & DeVillers\_reports                                         & 0.6123   & soccer      & 0.5734\\
        & \_sports         & 0.6527 & athletics                                                  & 0.6093   & baseball  & 0.5572\\
        & Sports         & 0.6479 & football                                                   & 0.5927   & football     & 0.5556\\
        & Sport          & 0.6198 & sporting\_events                                           & 0.5816   & espn   & 0.5110\\
        & \_athletic       & 0.6132 & soccer                                                     & 0.5805   & athletics      & 0.5071\\
        & \_athletes       & 0.6090 & al\_Sunaidy                                                & 0.5768   & athletic    & 0.5070\\
        & \_SPORTS         & 0.6086 & baseball                                                   & 0.5658   & entertainment    & 0.5062\\
        & \_football       & 0.6076 & limited  edition\_MGTF                                     & 0.5636   & hockey   & 0.4972\\
        & \_soccer         & 0.5956 & OSAA\_oversees                                             & 0.5610   & news  & 0.4953\\
        & \_basketball     & 0.5938 & motorsports                                                & 0.5515   & athletes   & 0.4897 \\
        & \_tennis         & 0.5873 & athletic                                                   & 0.5434   & golf    & 0.4781 \\
        & \_baseball       & 0.5846 & writers\_Jim\_Vertuno                                      & 0.5395   & tennis   & 0.4762 \\
\hline
science & \_Science        & 0.8053 & faith\_Jezierski                                           & 0.6965   & sciences     & 0.6844 \\
        & \_scientific     & 0.7044 & sciences                                                   & 0.6821   & physics      & 0.6518 \\
        & \_sciences       & 0.7001 & biology                                                    & 0.6776   & scientific   & 0.6487 \\
        & science        & 0.6901 & scientific                                                 & 0.6535   & biology      & 0.6283 \\
        & \_scientists     & 0.6895 & mathematics                                                & 0.6301   & mathematics  & 0.6216 \\
        & \_scientist      & 0.6889 & Hilal\_Khashan\_professor                                  & 0.6153   & research     & 0.6128 \\
        & \_physics        & 0.6700 & impeach\_USADA                                             & 0.6149   & technology   & 0.6056 \\
        & Science        & 0.6638 & professor\_Kent\_Redfield                                  & 0.6144   & fiction      & 0.5882 \\
        & \_biology        & 0.6482 & physics\_astronomy                                         & 0.6105   & professor    & 0.5873 \\
        & \_neuroscience   & 0.6223 & {bionic\_prosthetic\_fingers}                              & 0.6083   & chemistry    & 0.5856 \\
        & \_astronomy      & 0.6094 & {professor\_Burdett\_Loomis}                               & 0.6065   & university   & 0.5850 \\
        & \_mathematics    & 0.5957 & {Board\_BONU\_specialty}                                   & 0.6063   & engineering  & 0.5757 \\
        & \_scientifically & 0.5897 & Science                                                    & 0.6052   & psychology   & 0.5684 \\
        & \_Sciences       & 0.5796 & portal\_EurekAlert                                         & 0.5958   & institute    & 0.5678 \\
        & \_chemistry      & 0.5720 & Shlomo\_Avineri\_professor                                 & 0.5942   & literature   & 0.5656 \\
\hline
\end{tabularx}
    \caption{The 15 nearest neighbors of ``sports'' and ``science'' constructed from three word embeddings: LM, word2vec, and GLoVe, with their respective similarities to the corresponding core words.}
    \label{tab:E}
\end{table*}

\section{Instruction Format for Prompting \texttt{Mistral-7B-Instruct-v0.2}}
\label{app:mistral}

We use the prompts adapted from \cite{bach2022promptsource} for datasets in English and manually written prompt for datasets in French. We refer to \footnote{\url{https://www.promptingguide.ai/models/mistral-7b}} for prompt format.

For zero shot inference:
\begin{itemize}
    \item \textbf{AG}
    \begin{lstlisting}
        [INST]You are a topic labelling assistant. Given the following text:
        {text}
        Which topic is this text about among:
        world,
        sports,
        business,
        science/technology
        ?[/INST]
    \end{lstlisting}
    \item \textbf{Yahoo}
    \begin{lstlisting}
        [INST]You are a topic labelling assistant.
        {question_title} {question_content}
        Which topic is this question about? among:
        society & culture
        science & mathematics
        health
        education & reference
        computers & internet
        sports
        business & finance
        entertainment & music
        family & relationships
        politics & government
        ?[/INST]
    \end{lstlisting}
    \item \textbf{DBpedia}
    \begin{lstlisting}
        [INST]You are a text category annotator. Given the following text:
        {title}{content}
        Given a list of categories:
        company, 
        educational institution, 
        artist, 
        athlete, 
        office holder, 
        mean of transportation, 
        building, 
        natural place, 
        village, 
        animal, 
        plant, 
        album, 
        film, 
        written work. 
        Which category does this text belong to?[/INST]
    \end{lstlisting}
    \item \textbf{MLSUM Fr}
    \begin{lstlisting}
        [INST]Tu es un assistant de classification de thème. Lire le texte suivant:
        {title}
        {summary}
        Ce texte parle de quel thème parmi: 
        économie, 
        opinion, 
        politique, 
        société, 
        culture, 
        sport, 
        environnement, 
        technologie, 
        éducation, 
        justice 
        ?[/INST]
    \end{lstlisting}
    \item \textbf{FrN}
    \begin{lstlisting}
        [INST]Tu es un assistant de classification de secteur des articles de presse. Lire le texte suivant
        {title}
        {snippet}
        Ce texte appartient à quel secteur parmi: 
        aéronautique, 
        armement, 
        agroalimentaire, 
        automobile, 
        distribution - commerce, 
        électricité, 
        finance, 
        pétrole - gaz, 
        promotion immobilière, 
        tourisme - hôtellerie - restauration, 
        transport 
        ?[/INST]"
    \end{lstlisting}
\end{itemize}

For few-shot in-context learning, we insert the 32 demonstrations into the prompt.
\begin{itemize}
    \item \textbf{AG}
    \begin{lstlisting}
        [INST]You are a topic labeling assistant. Given a text, you need to answer which topic is this text about.
        Here are some examples:
        
        Text: {text_i}
        Label: {label_i}

        Which topic is this text about among:
        world
        sports
        business
        science/technology?
        
        Text: {text}
        Label: [/INST]
    \end{lstlisting}
    \item \textbf{Yahoo}
    \begin{lstlisting}
        [INST]You are a topic labeling assistant. Given a question, you need to answer which topic is this question about.
        Here are some examples:

        Text: {question_title_i} {question_content_i}
        Label: {topic_i}

        Which topic is this question about among:
        society & culture
        science & mathematics
        health
        education & reference
        computers & internet
        sports
        business & finance
        entertainment & music
        family & relationships
        politics & government
        ?
        
        Text: {question_title} {question_content}
        Label: [/INST]
    \end{lstlisting}
    \item \textbf{DBpedia}
    \begin{lstlisting}
        [INST]You are a categorizing assistant. Given a title and a description, you need to determine which category does the title belong to.
        Here are some examples:

        Title: {title_i}
        Description: {content_i}
        Label: {label_i}

        Which category does this belong to among:
        society & culture
        science & mathematics
        health
        education & reference
        computers & internet
        sports
        business & finance
        entertainment & music
        family & relationships
        politics & government
        ?
        
        Title: {title}
        Description: {content}
        Label: [/INST]
    \end{lstlisting}
    \item \textbf{MLSUM Fr}
    \begin{lstlisting}
        [INST]Tu es un assistant de classification de thème. Basé sur un titre et un texte, tu dois prédire le thème dont ce texte parle.
        Voici quelques exemples:

        Titre: {title_i}
        Texte: {summary_i}
        Thème: {label_i}

        Ce texte parle de quel thème parmi: 
        économie, 
        opinion, 
        politique, 
        société, 
        culture, 
        sport, 
        environnement, 
        technologie, 
        éducation, 
        justice
        ? 
        
        Titre: {title}
        Texte: {summary}
        Thème: [/INST]
    \end{lstlisting}
    \item \textbf{FrN}
    \begin{lstlisting}
        [INST]Tu es un assistant de classification de secteur des articles de presse. Basé sur un titre et un texte, tu dois prédire le secteur auquel ce texte appartient.
        Voici quelques exemples:

        Titre: {title_i}
        Texte: {snippet_i}
        Secteur: {sector_i}

        Ce texte appartient à quel secteur parmi: 
        aéronautique, 
        armement, 
        agroalimentaire, 
        automobile, 
        distribution - commerce, 
        électricité, 
        finance, 
        pétrole - gaz, 
        promotion immobilière, 
        tourisme - hôtellerie - restauration, 
        transport
        ? 
        
        Titre: {title}
        Texte: {snippet}
        Secteur: [/INST]
    \end{lstlisting}
\end{itemize}

\end{document}